\documentclass{article}
\usepackage{ijcai24}

\usepackage{times}
\usepackage{soul}
\usepackage{url}
\usepackage[hidelinks]{hyperref}
\usepackage[utf8]{inputenc}
\usepackage[small]{caption}
\usepackage{graphicx}
\usepackage{amsmath}
\usepackage{amsthm}
\usepackage{booktabs}
\usepackage{algorithm}
\usepackage{algorithmic}
\usepackage[switch]{lineno}
\usepackage{multirow}
\usepackage{array}
\usepackage{subcaption}
\usepackage[normalem]{ulem}

\title{A Survey of Large Language Models in Finance (FinLLMs)}

\author{
Jean Lee$^1$\and
Nicholas Stevens$^2$\and
Soyeon Caren Han$^{1,3}$\and
Minseok Song$^4$
\affiliations
$^1$The University of Sydney, Australia, 
$^2$NRS Technology, Australia \\
$^3$The University of Melbourne, Australia \\
$^4$Pohang University of Science and Technology (POSTECH), South Korea\\
\emails
\{jean.lee, caren.han\}@sydney.edu.au,
nick@nrs.technology, 
mssong@postech.ac.kr
}

\begin{document}

\maketitle

\begin{abstract}
Large Language Models (LLMs) have shown remarkable capabilities across a wide variety of Natural Language Processing (NLP) tasks and have attracted attention from multiple domains, including financial services. Despite the extensive research into general-domain LLMs, and their immense potential in finance, Financial LLM (FinLLM) research remains limited. This survey provides a comprehensive overview of FinLLMs, including their history, techniques, performance, and opportunities and challenges. Firstly, we present a chronological overview of general-domain Pre-trained Language Models (PLMs) through to current FinLLMs, including the GPT-series, selected open-source LLMs, and financial LMs. Secondly, we compare five techniques used across financial PLMs and FinLLMs, including training methods, training data, and fine-tuning methods. Thirdly, we summarize the performance evaluations of six benchmark tasks and datasets. In addition, we provide eight advanced financial NLP tasks and datasets for developing more sophisticated FinLLMs. Finally, we discuss the opportunities and the challenges facing FinLLMs, such as hallucination, privacy, and efficiency. To support AI research in finance, we compile a collection of accessible datasets and evaluation benchmarks on GitHub. \footnote{https://github.com/adlnlp/FinLLMs}  
\end{abstract}


\section{Introduction}
Research on Large Language Models (LLMs) has grown rapidly in both academia and industry, with notable attention to LLM applications such as ChatGPT. Inspired by Pre-trained Language Models (PLMs) \cite{devlin2018bert,radford2018improving}, LLMs are empowered by transfer learning and built upon the Transformer architecture \cite{vaswani2017attention}  using large-scale textual corpora. Researchers have discovered that scaling models \cite{radford2019language} to sufficient sizes not only enhances model capacity but also enables surprising emergent properties, such as in-context learning \cite{brown2020language}, that do not appear in small-scale language models. 
Language Models (LMs) can be categorized based on parameter size, and the research community has created the term “Large Language Models (LLM)” for PLMs of substantial size, typically exceeding 7 billion parameters \cite{zhao2023survey}. The technical evolution of LLMs has resulted in a remarkable level of homogenization \cite{bommasani2021opportunities}, meaning that a single model could yield strong performance across a wide range of tasks. The capability of LLMs has facilitated the adaptation of various forms of multimodal data (e.g., text, image, audio, video, and tabular data) and multimodal models across AI and interdisciplinary research communities.  

In the financial domain, there has been growing interest in applying NLP across various financial tasks, including sentiment analysis, question answering, and stock market prediction. The rapid advancement of general-domain LLMs has spurred an investigation into Financial LLMs (FinLLMs), employing methods such as mixed-domain LLMs with prompt engineering and instruction fine-tuned LLMs with prompt engineering. 
While general LLMs are extensively researched and reviewed \cite{zhao2023survey,yang2023harnessing,zhang2023instruction}, the field of financial LLMs \cite{li2023large} is at an early stage. Considering the immense potential of LLMs in finance, this survey provides a holistic overview of FinLLMs and discusses future directions that can stimulate interdisciplinary studies. We acknowledge that this research focuses on LMs in English. The key contributions of this survey paper are summarized below.

\begin{itemize}
\item To the best of our knowledge, this is the first comprehensive survey of FinLLMs that explores the evolution from general-domain LMs to financial-domain LMs.
\item We compare five techniques used across four financial PLMs and four financial LLMs, including training methods and data, and instruction fine-tuning methods. 
\item We summarize the performance evaluation of six benchmark tasks and datasets between different models, and provide eight advanced financial NLP tasks and datasets for the development of advanced FinLLMs. 
\item We discuss the opportunities and the challenges of FinLLMs, with regard to datasets, techniques, evaluation, implementation, and real-world applications.

\end{itemize}

\begin{figure*}
	\centering
	\includegraphics[width=\linewidth]{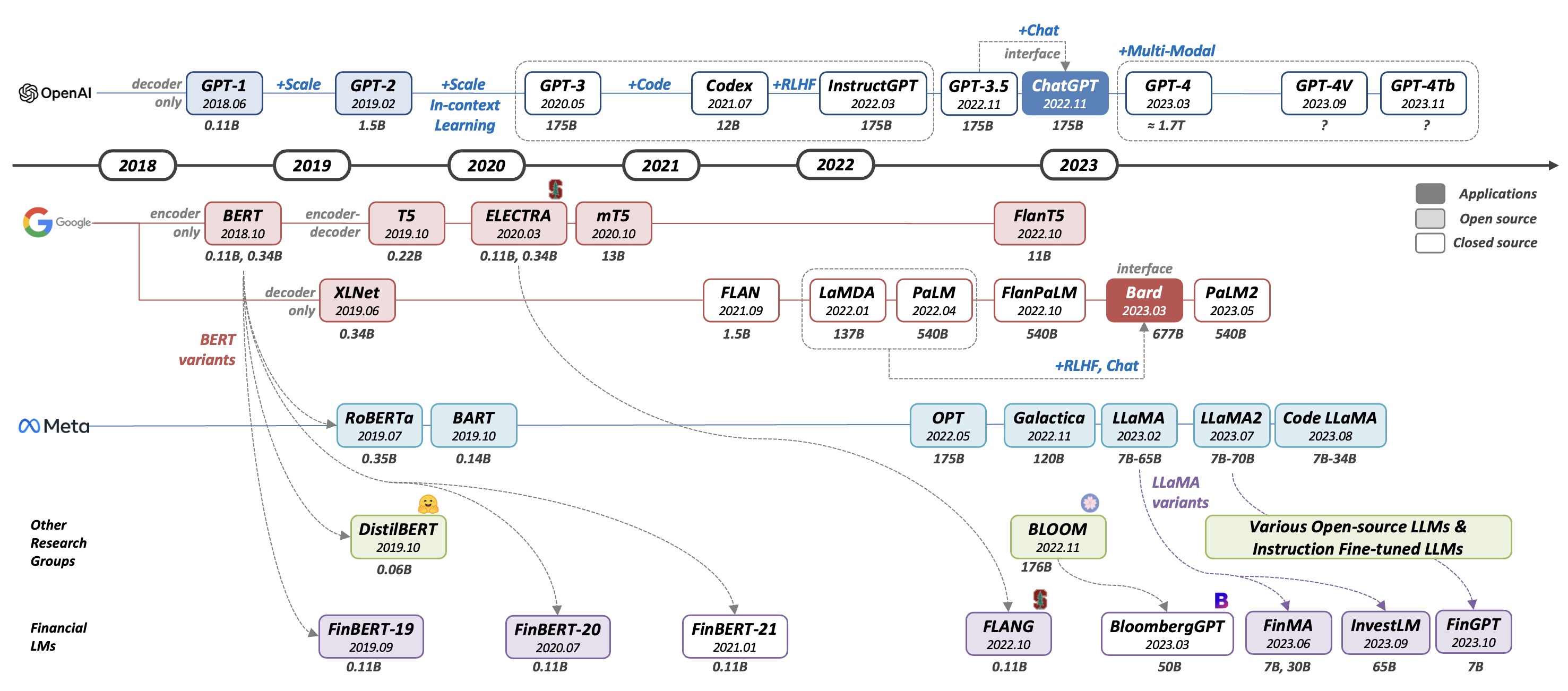}
	\caption{Timeline showing the evolution of selected PLM/LLM releases from the general domain to the financial domain.}
	\label{LLM_timeline}
\end{figure*}

\section{Evolution Trends: from General to Finance}


\subsection{General-domain LMs}
Since the introduction of the Transformer \cite{vaswani2017attention} architecture by Google in 2017,  Language Models (LMs) are generally pre-trained with either \textbf{discriminative} or \textbf{generative} objectives. Discriminative pre-training uses a masked language model to predict the next sentence and features an encoder-only or an encoder-decoder architecture. Generative pre-training uses autoregressive language modeling to predict the next token and features a decoder-only architecture.
Figure \ref{LLM_timeline} illustrates the evolutionary timeline from general-domain LMs to financial-domain LMs.

\subsubsection{GPT-Series}
The Generative Pre-trained Transformer (GPT) series of models started with \textbf{GPT-1} (110M) \cite{radford2018improving}. Since then, the OpenAI team focused on scaling the model, and \textbf{GPT-2} (1.5B) \cite{radford2019language} was released in 2019. GPT-2 identified the power of scaling and a probabilistic approach for multi-task problem-solving. In 2020, \textbf{GPT-3} with 175B parameters was released \cite{brown2020language}. This was a significant milestone for LLMs, as it introduced an emergent capability of LLMs; in-context learning. \textbf{In-context learning} refers to the model acquiring capabilities that were not explicitly trained, allowing language models to understand human language and produce outcomes beyond their original pre-training objectives.

Ongoing efforts to improve LLMs have resulted in the introduction of \textbf{ChatGPT}, in November 2022. This application combines GPT-3 (In-context learning), Codex (LLMs for code), and InstructGPT (Reinforcement Learning with Human Feedback, RLHF).
The success of ChatGPT has led to further development of significantly larger models, including \textbf{GPT-4} (estimated 1.7T parameters). GPT-4 demonstrates human-level performance, capable of passing law and medical exams, and handling multimodal data.

OpenAI continues to build extremely large language models, aiming to enhance the model's capabilities in handling multimodal data, as well as providing APIs for the development of real-world applications. Despite the mainstream popularity and adoption, real-world applications in finance utilizing their APIs have not yet been fully explored.


\begin{figure*}
	\centering
	\includegraphics[width=\linewidth]{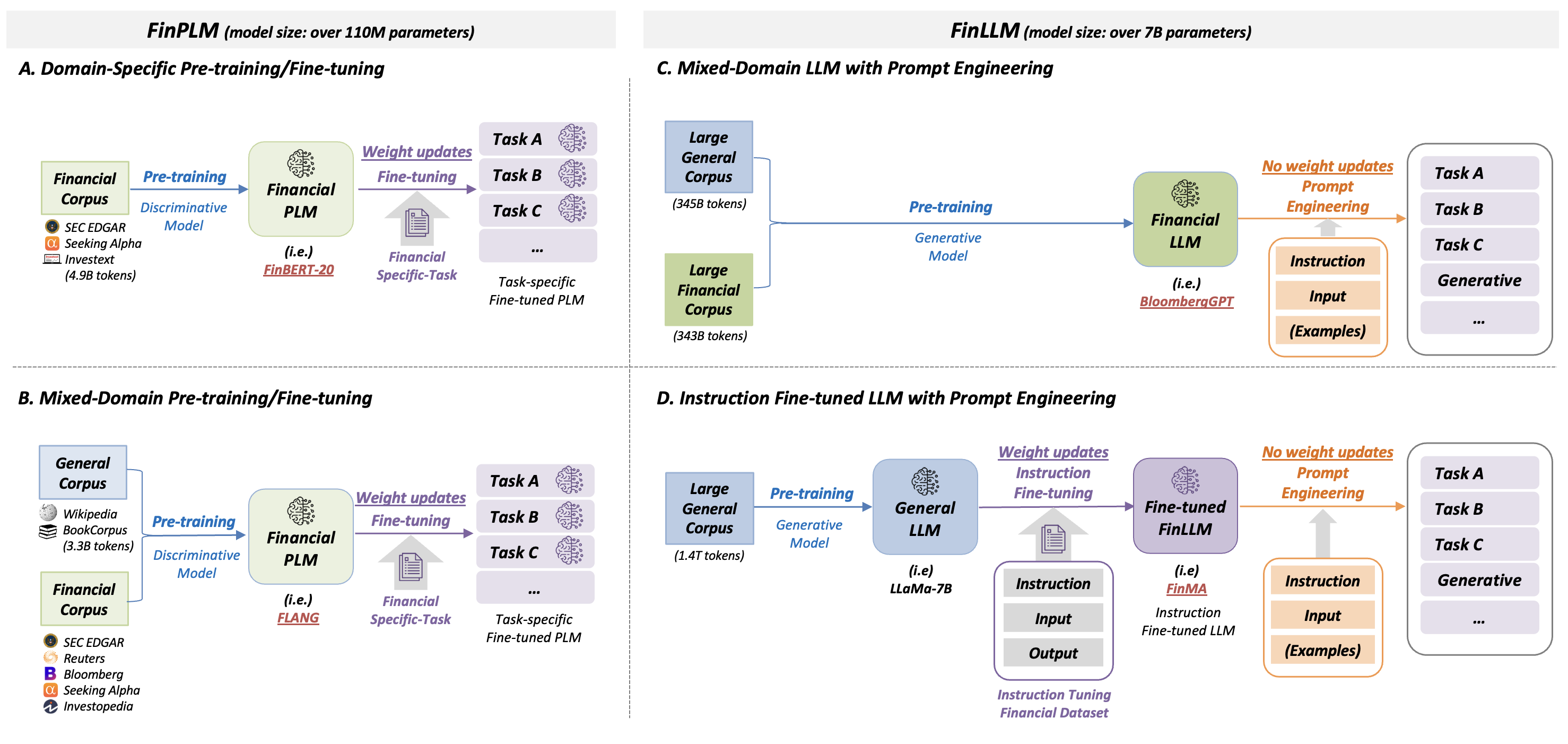}
	\caption{Comparison of techniques used in financial LMs: from FinPLMs to FinLLMs.}
	\label{techniques}
\end{figure*}

\subsubsection{Open-source LLMs}
Prior to the era of LLMs, the research community often released open-source PLMs such as Bidirectional Encoder Representations from Transformers (\textbf{BERT}, base-110M parameters) \cite{devlin2018bert}. BERT is the foundational model for many early PLMs, including FinBERT. Since OpenAI shifted from open-source to closed-source LLMs, the trend across LLM research is a reduction in the release of open-source models. However, in February 2023, Meta AI released the open-source LLM, \textbf{LLaMA} (7B, 13B, 33B, 65B parameters) \cite{touvron2023llama}, and this encouraged the development of diverse LLMs using LLaMA. Similar to BERT variants, LLaMA variants quickly proliferated by adopting various techniques such as Instruction Fine-Tuning (IFT) \cite{zhang2023instruction} and Chain-of-Thought (CoT) Prompting \cite{wei2022chain}. 


There have also been significant efforts by the research community to generate open-source LLMs to reduce the reliance on corporate research and proprietary models. \textbf{BLOOM} (176B) \cite{scao2022bloom} was built by a collaboration of hundreds of researchers from the BigScience Workshop. This open-source LLM was trained on 46 natural languages and 13 programming languages.

\subsection{Financial-domain LMs}
Domain-specific LMs, such as financial-domain LMs, are commonly built using general-domain LMs. In finance, there are primarily four financial PLMs (FinPLMs) and four financial LLMs (FinLLMs). Within the four FinPLMs, FinBERT-19 \cite{araci2019finbert}, FinBERT-20  \cite{yang2020finbert}, and FinBERT-21 \cite{liu2021finbert} are all based on BERT, while FLANG \cite{shah2022flue} is based on ELECTRA \cite{clark2020electra}. Within the four FinLLMs, FinMA \cite{xie2023pixiu}, InvestLM \cite{yang2023investlm}, and FinGPT \cite{wang2023fingpt} are based on LLaMA or other open-source-based models, while BloombergGPT \cite{wu2023bloomberggpt} is a BLOOM-style closed-source model.


\section{Techniques: from FinPLMs to FinLLMs} \label{sec3}
While our survey focuses on FinLLMs, it is important to acknowledge that previous studies on FinPLMs as they formed the groundwork for FinLLM development. We reviewed three techniques used by the four FinPLMs and two techniques used by the four FinLLMs. Figure \ref{techniques} illustrates technical comparisons of building financial LMs \footnote{The continual pre-training diagram can be found on our GitHub.}, and Table \ref{tab:FinLLMs} shows a summary of FinPLMs/FinLLMs including pre-training techniques, fine-tuning, and evaluation. 




\subsection {Continual Pre-training}
Continual pre-training of LMs aims to train an existing general LM with new domain-specific data on an incremental sequence of tasks \cite{ke2022continual}. 

\textbf{FinBERT-19} \cite{araci2019finbert} \footnote{https://huggingface.co/ProsusAI/finbert} is the first FinBERT model released for financial sentiment analysis and implements three steps: 1) the initialization of the general-domain BERT PLM (3.3B tokens), 2) continual pre-training on a financial-domain corpus, and 3) fine-tuning on financial domain-specific NLP tasks. The fine-tuned financial LM is released on HuggingFace, and this FinBERT-19 is a task-dependent model for the financial sentiment analysis task.

\subsection {Domain-Specific Pre-training from Scratch}
The domain-specific pre-training approach involves training a model exclusively on an unlabeled domain-specific corpus while following the original architecture and its training objective.

\textbf{FinBERT-20} \cite{yang2020finbert} \footnote{https://github.com/yya518/FinBERT} is a finance domain-specific BERT model, pre-trained on a financial communication corpus (4.9B tokens). 
The author released not only the FinBERT model but also FinVocab uncased/cased, which has a similar token size to the original BERT model. FinBERT-20 also conducted a sentiment analysis task for fine-tuning experiments on the same dataset of FinBERT-19. 

\begin{table*}[]
  \centering
  \scriptsize
  \setlength{\tabcolsep}{0.9mm}{
  \renewcommand{\arraystretch}{1.3}

  \newcolumntype{L}{>{\fontfamily{ptm}\selectfont}l}
  \newcolumntype{C}{>{\fontfamily{ptm}\selectfont}c}
  \newcolumntype{R}{>{\fontfamily{ptm}\selectfont}r}
  
\begin{tabular}{LLLCLLLLCCCCC}
\toprule
\multicolumn{5}{L}{\textbf{}} &
  \multicolumn{1}{L}{\textbf{PT}} &
  \multicolumn{2}{L}{\textbf{Evaluation}} &
  \multicolumn{3}{C}{\textbf{Open Source}} &
  \textbf{} \\ 
  \cline{6-11}
\textbf{Category} &
  \textbf{Model} &
  \textbf{Backbone} &
  \textbf{Paras.} &
  \textbf{Techniques} &
  \textbf{PT Data Size} &
  \textbf{Task} &
  \textbf{Dataset} &
  \textbf{Model} &
  \textbf{PT} &
  \textbf{IFT} &
  \textbf{Venue} \\
  \midrule
\multirow{1}{*}{\begin{tabular}[C]{@{}l@{}}FinPLM\\ (Disc.)\end{tabular}} 
&
  \begin{tabular}[c]{@{}l@{}} FinBERT-19 \\ \cite{araci2019finbert} \end{tabular}  &
  BERT &
  0.11B &
  Post-PT, FT &
  \begin{tabular}[c]{@{}l@{}}(G) 3.3B words \\ (F) 29M words\end{tabular} &
  [SA] &
  FPB, FiQA-SA &
  {\href{https://huggingface.co/ProsusAI/finbert}{Y}} &
  N &
  N &
  \begin{tabular}[c]{@{}c@{}}ArXiv \\ Aug 2019\end{tabular} \\
\cline{2-12}  
 &
  \begin{tabular}[c]{@{}l@{}} FinBERT-20 \\ \cite{yang2020finbert} \end{tabular} &
  BERT &
  0.11B &
  PT, FT &
  (F) 4.9B tokens &
  [SA] &
  FPB, FiQA-SA, AnalystTone &
  {\href{https://github.com/yya518/FinBERT}{Y}} &
  Y &
  N &
  \begin{tabular}[c]{@{}c@{}}ArXiv\\ Jul 2020\end{tabular} \\
\cline{2-12} 
 &
  \begin{tabular}[c]{@{}l@{}} FinBERT-21 \\ \cite{liu2021finbert} \end{tabular} & 
  BERT &
  0.11B &
  PT, FT &
  \begin{tabular}[c]{@{}l@{}}(G) 3.3B words \\ (F) 12B words\end{tabular} &
  \begin{tabular}[c]{@{}l@{}}{[}SA{]}, {[}QA{]} \\ {[}SBD{]}\end{tabular} &
  \begin{tabular}[c]{@{}l@{}}FPB, FiQA-SA, FiQA-QA \\ FinSBD19\end{tabular} &
  N &
  N &
  N &
  \begin{tabular}[c]{@{}c@{}}IJCAI (S)\\ Jan 2021\end{tabular} \\
\cline{2-12} 
&
  \begin{tabular}[c]{@{}l@{}} FLANG \\ \cite{shah2022flue} \end{tabular} &
  ELECTRA  &
  0.11B &
  PT, FT &
  \begin{tabular}[c]{@{}l@{}}(G) 3.3B words \\ (F) 696k docs\end{tabular} &
  \begin{tabular}[c]{@{}l@{}}{[}SA{]}, {[}TC{]}\\ {[}NER{]}, {[}QA{]}, {[}SBD{]}\end{tabular} &
  \begin{tabular}[c]{@{}l@{}}FPB, FiQA-SA, Headline \\ FIN, FiQA-QA, FinSBD21\end{tabular} &
  {\href{https://github.com/SALT-NLP/FLANG}{Y}} &
  Y &
  N &
  \begin{tabular}[c]{@{}c@{}}EMNLP\\ Oct 2022\end{tabular} \\
\midrule
\multirow{1}{*}{\begin{tabular}[c]{@{}l@{}}FinLLM\\ (Gen.)\end{tabular}}
&
  \begin{tabular}[c]{@{}l@{}} BloombergGPT \\ \cite{wu2023bloomberggpt} \end{tabular} & 
  BLOOM &
  50B &
  PT, PE &
  \begin{tabular}[c]{@{}l@{}}(G) 345B tokens \\ (F) 363B tokens\end{tabular} &
  \begin{tabular}[c]{@{}l@{}}{[}SA{]}, {[}TC{]} \\ {[}NER{]}, {[}QA{]} \end{tabular} &
  \begin{tabular}[c]{@{}l@{}}FPB, FiQA-SA, Headline \\ FIN, ConvFinQA\end{tabular} &
  N &
  N &
  N &
  \begin{tabular}[c]{@{}c@{}}ArXiv\\ Mar 2023\end{tabular} \\
\cline{2-12} 
 &
  \begin{tabular}[c]{@{}l@{}} FinMA \\ \cite{xie2023pixiu} \end{tabular} &
  LLaMA &
  7B, 30B &
  IFT, PE &
  (G) 1T tokens &
  \begin{tabular}[c]{@{}l@{}}{[}SA{]}, {[}TC{]}, \\ {[}NER{]}, {[}QA{]} \\ {[}SMP{]}\end{tabular} &
  \begin{tabular}[c]{@{}l@{}}FPB, FiQA-SA, Headline \\ FIN, FinQA, ConvFinQA, \\ StockNet, CIKM18, BigData22 \end{tabular} &
  {\href{https://github.com/chancefocus/PIXIU}{Y}} &
  Y &
  Y &
  \begin{tabular}[c]{@{}c@{}}NIPS (D)\\ Jun 2023\end{tabular} \\
  \cline{2-12} 
 &
  \begin{tabular}[c]{@{}l@{}} InvestLM \\ \cite{yang2023investlm} \end{tabular} & 
  LLaMA &
  65B &
  \begin{tabular}[c]{@{}l@{}}IFT, PE \\ PEFT \end{tabular} &
  (G) 1.4T tokens &
  \begin{tabular}[c]{@{}l@{}}{[}SA{]}, {[}TC{]} \\ {[}QA{]}, {[}Summ{]} \end{tabular} &
  \begin{tabular}[c]{@{}l@{}}FPB, FiQA-SA, FOMC \\ FinQA, ECTSum\end{tabular} &
  {\href{https://github.com/AbaciNLP/InvestLM}{Y}} &
  N &
  N &
  \begin{tabular}[c]{@{}c@{}}ArXiv\\ Sep 2023\end{tabular} \\
  \cline{2-12} 
&
  \begin{tabular}[c]{@{}l@{}} FinGPT \\ \cite{wang2023fingpt} \end{tabular} & 
  \begin{tabular}[c]{@{}l@{}}6 open-source \\ LLMs \end{tabular} &
  7B &
  \begin{tabular}[c]{@{}l@{}}IFT, PE \\ PEFT \end{tabular} & 
  \begin{tabular}[c]{@{}l@{}}(G) 2T tokens \\ (e.g. LLaMA2)\end{tabular} &
  \begin{tabular}[c]{@{}l@{}}{[}SA{]}, {[}TC{]} \\ {[}NER{]}, {[}RE{]} \end{tabular} &
  \begin{tabular}[c]{@{}l@{}}FPB, FiQA-SA, Headline \\ FIN, FinRED\end{tabular} &
  {\href{https://github.com/AI4Finance-Foundation/FinGPT}{Y}} &
  Y &
  Y &
  \begin{tabular}[c]{@{}c@{}}NIPS (W) \\Oct 2023\end{tabular}\\
\bottomrule
\end{tabular}
}
  \caption{\fontfamily{ptm}\selectfont A Summary of FinPLMs and FinLLMs. The abbreviations correspond to
  Paras.= Model Parameter Size (Billions); Disc. = Discriminative, Gen. = Generative; Post-PT = Post-Pre-training, PT = Pre-training, FT = Fine-Tuning, PE = Prompt Engineering, IFT = Instruction Fine-Tuning, PEFT = Parameter Efficient Fine-Tuning; (G) = General domain, (F) = Financial domain; (in Evaluation) [SA] Sentiment Analysis, [TC] Text Classification, [SBD] Structure Boundary Detection, [NER] Named Entity Recognition, [QA] Question Answering, [SMP] Stock Movement Prediction, [Summ] Text Summarization, [RE] Relation Extraction; (in Venue) (S) = Special Track, (D) = Datasets and Benchmarks Track, (W) = Workshop. In open source, it is marked as Y if it is publicly accessible as of Dec 2023.}
  \label{tab:FinLLMs}
\end{table*}

\subsection {Mixed-Domain Pre-training}
The mixed-domain pre-training approach involves training a model using both a general-domain corpus and a domain-specific corpus. 
The assumption is that general-domain text remains relevant, while the financial domain data provides knowledge and adaptation during the pre-training process.

\textbf{FinBERT-21} \cite{liu2021finbert} \footnote{As of Dec 2023, the link mentioned in the paper does not exist.} is another BERT-based PLM designed for financial text mining, trained simultaneously on a general corpus and a financial domain corpus.
FinBERT-21 employs multi-task learning across six self-supervised pre-training tasks, enabling it to efficiently capture language knowledge and semantic information. 
FinBERT-21 conducted experiments on Sentiment Analysis as well as provided experiment results for two additional tasks; Sentence Boundary Detection and Question Answering.

\textbf{FLANG} \cite{shah2022flue} \footnote{https://github.com/SALT-NLP/FLANG} is a domain-specific model using financial keywords and phrases for masking, and follows the training strategy of ELECTRA \cite{clark2020electra}. 
This research first introduces Financial Language Understanding Evaluation (\textbf{FLUE}), a collection of five financial NLP benchmark tasks. The tasks include Sentiment Analysis, Headline Text Classification, Named Entity Recognition, Structure Boundary Detection, and Question Answering. 

\subsection {Mixed-Domain LLM with Prompt Engineering}
Mixed-domain LLMs are trained on both a large general corpus and a large domain-specific corpus. Then, users describe the task and optionally provide a set of examples in human language. This technique is called Prompt Engineering and uses the same frozen LLM for several downstream tasks with no weight updates.
This survey does not explore prompt engineering but instead references recent surveys \cite{liu2023pre}. 

\textbf{BloombergGPT} \cite{wu2023bloomberggpt} \footnote{This model and its associated data are closed-source.} is the first FinLLM that utilizes the BLOOM model \cite{scao2022bloom}. It is trained on a large general corpus (345B tokens), and a large financial corpus (363B tokens). The financial corpus, FinPile, contains data collected from the web, news, filings, press, and Bloomberg’s proprietary data. The authors conducted financial NLP tasks (5 benchmark tasks and 12 internal tasks) as well as 42 general-purpose NLP tasks.

\subsection {Instruction Fine-tuned LLM with Prompt Engineering}
Instruction tuning is the additional training of LLMs using explicit text instructions to enhance the capabilities and controllability of LLMs. Research on instruction tuning can be classified into two main areas \cite{zhang2023instruction}: 1) the construction of instruction datasets, and 2) the generation of fine-tuned LLMs using these instruction datasets. 
In finance, researchers have started transforming existing financial datasets into instruction datasets and subsequently using these datasets for fine-tuning LLMs.


\textbf{FinMA} (or PIXIU) \cite{xie2023pixiu} \footnote{https://github.com/chancefocus/PIXIU} consists of two fine-tuned LLaMA models (7B and 30B) \cite{touvron2023llama} that use financial instruction datasets for financial tasks. It is constructed from a large-scale multi-task instruction dataset called Financial Instruction Tuning (FIT, 136k samples) by collecting nine publicly released financial datasets used across five different tasks. In addition to the five FLUE benchmark tasks, it includes the Stock Movement Prediction task. 

\textbf{InvestLM} \cite{yang2023investlm} \footnote{The instruction dataset is unavailable, but the model has been released on GitHub. https://github.com/AbaciNLP/InvestLM} is a fine-tuned LLaMA-65B model using a manually curated financial domain instruction dataset. The dataset includes Chartered Financial Analyst (CFA) exam questions, SEC filings, Stackexchange quantitative finance discussions, and Financial NLP tasks. The downstream tasks are similar to FinMA but also include a financial text Summarization task.


\textbf{FinGPT} \cite{yang2023fingpt} \footnote{https://github.com/AI4Finance-Foundation/FinGPT} is an open-sourced and data-centric framework, which provides a suite of APIs for financial data sources, an instruction dataset for financial tasks, and several fine-tuned financial LLMs. The FinGPT team has released several similar papers that describe the framework and an experiment paper \cite{wang2023fingpt} on the instruction fine-tuned FinLLMs using six open-source LLMs with the Low-Rank Adaptation (LoRA) \cite{hu2021lora} method. 


\section{Evaluation: Benchmark Tasks and Datasets} \label{eval}
As LLMs gain significant attention, evaluating them becomes increasingly critical. We summarize six financial NLP benchmark tasks and datasets, and review the evaluation results of models including FinPLMs, FinLLMs, ChatGPT, GPT-4, and task-specific State-of-the-Art (SOTA) models. The results \footnote{Figure and tables of each task can be found on our Github.} are referenced from original research or analysis research \cite{li2023chatgpt}, and SOTA results from task-specific models. 


\subsection {Sentiment Analysis (SA)}
The Sentiment Analysis (SA) task aims to analyze sentiment information from input text, including financial news and microblog posts. Most FinPLMs and FinLLMs report the evaluation results of this task using the Financial PhraseBank (FPB) and the FiQA SA dataset. 
The \textbf{FPB} \cite{malo2014good} dataset consists of 4,845 
English financial news articles. 
Domain experts annotated each sentence with one of three sentiment labels: Positive, Negative, or Neutral. The \textbf{FiQA-SA} \cite{maia201818} dataset 
consists of 1,173 posts from both headlines and microblogs. The sentiment scores are on a scale of [-1, 1],
and recent studies have converted this score into a classification task. 
Overall, FLANG-ELECTRA achieved the best results (92\% on F1) while FinMA-30B and GPT-4 achieved similar results (87\% on F1) with a 5-shot prompting. It suggests a practical approach for less complex tasks in terms of efficiency and costs. 

For further evaluation of SA, we include two open-released datasets: SemEval-2017 (Task 5) and StockEmotions. 
The \textbf{SemEval-2017} \cite{cortis2017semeval} dataset comprises 4,157 sentences collected from both headlines and microblogs. Similar to FiQA SA, the sentiment scores are on a scale of [-1, 1]. The \textbf{StockEmotions} \cite{lee2023stockemotions} dataset consists of 10,000 sentences collected 
microblogs that annotate binary sentiment and 12 fine-grained emotion classes that span the multi-dimensional range of investor emotions.

\subsection {Text Classification (TC)}
Text Classification (TC) is the task of classifying a given text or document into predefined labels based on its content. In financial text, there are often multiple dimensions of information beyond sentiment such as price directions or interest rate directions. FLUE includes the gold news \textbf{Headline} \cite{sinha2021impact} dataset for text classification. This dataset comprises 11,412 news headlines, labeled with a binary classification across nine labels such as “price up”, or “price down”.
Similar to the SA task, FLANG-ELECTRA and FinMA-30B with a 5-shot prompting achieved the best results (98\% on Avg. F1) and the performance of BERT and FinBERT-20 was also noteworthy (97\% on Avg. F1). 

As TC is a broad task depending on the dataset and its predefined labels, we include three open-released financial TC datasets for further research: FedNLP, FOMC, and Banking77.
The \textbf{FedNLP} \cite{lee2021fednlp} dataset comprises documents sourced from various Federal Open Market Committee (FOMC) materials.
The dataset is annotated with labels as Up, Maintain, or Down based on the Federal Reserve's Federal Funds Rate decision for the subsequent period. Similarly, the \textbf{FOMC} \cite{shah2023trillion} dataset is a collection of FOMC documents with the labels as Dovish, Hawkish, or Neutral, reflecting the prevailing sentiment conveyed within the FOMC materials. The \textbf{Banking77} \cite{casanueva2020efficient} dataset comprises 13,083 samples covering 77 intents related to 
banking customer service queries, such as “card loss” or “linking to an existing card”. This dataset is designed for intent detection and developing conversation systems.

\subsection {Named Entity Recognition (NER)}
The Named Entity Recognition (NER) task is the extraction of information from unstructured text and categorizing it into predefined named entities such as locations (LOC), organizations (ORG), and persons (PER).
For the financial NER task, the \textbf{FIN} dataset \cite{alvarado2015domain} is included in FLUE benchmarks. The FIN dataset comprises eight financial loan agreements sourced from the US Security and Exchange Commission (SEC) for credit risk assessment. 
GPT-4 with a 5-shot prompting (83\% on Entity F1) and FLANG-ELECTRA demonstrate notable performance (82\% on Entity F1), while other FinLLMs exhibit suboptimal results (61\%-69\% on Entity F1). 

For further research, we include a financial NER dataset, \textbf{FiNER-139} \cite{loukas2022finer}, consisting of 1.1M sentences annotated with 139 eXtensive Business Reporting Language (XBRL) word-level tags, sourced from the SEC. 
This dataset is designed for Entity Extraction and Numerical Reasoning tasks, predicting the XBRL tags (e.g., cash and cash equivalents) based on numeric input data within sentences (e.g., “24.8” million). 

\subsection {Question Answering (QA)}
Question Answering (QA) is a task to retrieve or generate answers to questions from an unstructured collection of documents. Financial QA is more challenging than general QA as it requires numerical reasoning across multiple formats. 
\textbf{FiQA-QA} \cite{maia201818} is for opinion-based QA, representing an early Financial QA dataset. 

Over time, Financial QA has evolved to include complex numerical reasoning in multi-turn conversations. This evolution involves the introduction of hybrid QA, which is to create paths to connect hybrid contexts including both tabular and textual content. 
\textbf{FinQA} \cite{chen2021finqa} is a single-turn hybrid QA dataset having 8,281 QA pairs and annotated by experts from the annual reports of S\&P 500 companies. 
\textbf{ConvFinQA} \cite{chen2022convfinqa}, an extension of FinQA, is a multi-turn conversational hybrid QA dataset, consisting of 3,892 conversations with 14,115 questions. 
Instead of using the FiQA-QA dataset, all FinLLMs conducted experiments on the FinQA and/or ConvFinQA datasets to assess their numerical reasoning capabilities. 
GPT-4 with a zero-shot prompting outperforms all other models (69\%-76\% on EM Accuracy), approaching the performance of human experts (Avg. 90\% on EM Accuracy). BloombergGPT's result (43\% on EM Accuracy) was slightly below the general crowd (47\% on EM Accuracy). 



\subsection {Stock Movement Prediction (SMP)}
The Stock Movement Prediction (SMP) task aims to predict the next day's price movement (e.g., up or down) based on historical prices and associated text data. As it requires the integration of time series problems with temporal dependencies from text information, it presents a complex task, where text data can act both as noise and signal. 
FinMA includes the SMP tasks for the first time, conducting experiments on three datasets; StockNet, CIKM18, and BigData22. 

\textbf{StockNet} \cite{xu2018stock} collected historical price data and Twitter data between 2014 and 2016 for 88 stocks listed in the S\&P, and is widely used for SMP tasks. The task is framed as a binary classification with a threshold: a price movement higher than 0.55\% is labeled as a rise (denoted as 1), while a movement less than -0.5\% is labeled as a fall (denoted as 0). 
Similarly, \textbf{CIKM18} \cite{wu2018hybrid} utilizes historical price and Twitter data in 2017 for 47 stocks in the S\&P 500. 
\textbf{BigData22} \cite{soun2022accurate} compiled data between 2019 and 2020 for 50 stocks in the US stock markets. Like StockNet, it adopts a binary classification formulation with a threshold. 
On average across these three datasets, GPT-4 with a zero-shot prompting achieves higher performance (54\% on Accuracy) than FinMA (52\% on Accuracy) and slightly lower results than the SOTA model (58\% on Accuracy). Although NLP metrics such as Accuracy are commonly used, these are insufficient for the SMP evaluation. It is important to consider financial evaluation metrics, such as the Sharpe ratio, as well as backtesting simulation results. 

\subsection {Text Summarization (Summ)}
Summarization (Summ) is the generation of a concise summary from documents while conveying its key information via either an extractive or an abstractive approach. In finance, it has been relatively underexplored due to the lack of benchmark datasets, challenges with domain experts' evaluations, and the need for disclaimers when presenting financial advice. 
InvestLM includes summarization tasks for the first time, conducting experiments on the ECTSum dataset. \textbf{ECTSum} \cite{mukherjee2022ectsum} consists of 2,425 document-summary pairs, containing Earnings Call Transcripts (ECTs) and bullet-point summarizations from Reuters.
It reports evaluation results on various metrics, including ROUGE-1, ROUGE-2, ROUGE-L, and BERTScore.
Similar to other complex financial tasks, the task-specific SOTA model (47\% on ROUGE-1) outperforms all LLMs. According to the authors of InvestLM, while GPT-4 with a zero-shot prompting (30\% on ROUGE-1) shows superior performance compared to FinLLMs, the commercial models generate decisive answers. 

The summarization task offers significant development opportunities, exploring whether FinLLMs can outperform task-specific SOTA models. For ongoing research, we include the financial summarization dataset, \textbf{MultiLing 2019} \cite{el2019multiling}, containing 3,863 document-summary pairs extracted from UK annual reports listed on the London Stock Exchange (LSE). It provides at least two gold-standard summaries for each annual report.

\subsection {Discussion}
Within the six benchmarks, the performance of mixed-domain FinPLMs is noteworthy for the SA, TC, and NER tasks, suggesting that using a PLM with fine-tuning for a specific task can be a practical approach depending on the task complexity. For QA, SMP, and Summ tasks, the task-specific SOTA models outperform all LLMs, indicating areas for improvement in FinLLMs. Notably, GPT-4 shows impressive performance across all benchmarks except the Summ task, indicating that scaling models alone may not be adequate for optimal performance in finance. As most instruction-finetuned FinLLMs used the same datasets for their evaluation, we include additional datasets for future research. 

\section{Advanced Financial NLP Tasks and Datasets}
Properly designed benchmark tasks and datasets are a crucial resource to assess the capability of LLMs, however, the current 6 benchmark tasks have yet to address more complex financial NLP tasks. In this section, we present 8 advanced benchmark tasks and compile associated datasets for each. 


The \textbf{Relation Extraction} (RE) task aims to identify and classify relationships between entities implied in text. Similar to NER, this task is part of Information Extraction. The \textbf{FinRED} \cite{sharma2022finred} dataset is released for RE and is curated from financial news and earning call transcripts, containing 29 relation tags (e.g. owned by) specific to the finance domain.

\textbf{Event Detection} (ED) in finance involves identifying the impact of how investors perceive and assess related companies. The \textbf{Event-Driven Trading} (EDT) \cite{zhou2021trade} dataset is released for ED and includes 11 types of corporate event detection. EDT comprises 9,721 news articles with token-level event labels, and an additional 303,893 news articles with minute-level timestamps and stock price labels.

\textbf{Causality Detection} (CD) in finance seeks to identify cause-and-effect relationships within factual text, aiming to develop an ability to generate meaningful financial narrative summaries. The Workshop on Financial Narrative Processing (FNP) addresses this task every year and contributes datasets. One of the open-released dataset from FNP, \textbf{FinCausal20} \cite{mariko2020financial} shares two tasks: detecting a causal scheme in a given text and identifying cause-and-effect sentences.

\textbf{Numerical Reasoning} (NR) in finance aims to identify numbers and mathematical operators in either digit or word form, in order to perform calculations and comprehend financial context (e.g. cash and cash equivalent). Some datasets introduced for NER and QA tasks are also designed for numerical reasoning, including: \textbf{FiNER-139} \cite{loukas2022finer}, \textbf{FinQA} \cite{chen2021finqa}, \textbf{ConvFinQA} \cite{chen2022convfinqa}. 

\textbf{Structure Recognition} (SR) is a task focused on Structure Boundary Detection within a document (e.g. text, tables, or figures) and recognising the logical relationships between tables and surrounding content, or between cells within a table. IBM Research has released the \textbf{FinTabNet} \cite{zheng2021global} dataset, collected from earnings reports of S\&P 500 companies. This dataset comprises unstructured PDF documents with detailed annotations of table structures. The FinQA and ConvFinQA datasets, included in QA tasks, have been further developed from FinTabNet.

\textbf{Multimodal} (MM) understanding is a challenging task across many domains. Recently, several multimodal financial datasets have been introduced. \textbf{MAEC} \cite{li2020maec} compiles the multimodal data (text, time series, and audio) from earnings call transcripts on a larger scale, with 3,443 instances and 394,277 sentences. Additionally, \textbf{MONOPOLY} \cite{mathur2022monopoly} introduces video data from monetary policy call transcripts across six central banks, sharing 24,180 samples from 340 videos with text scripts and time series.

\textbf{Machine Translation} (MT) in finance aims to not only translate sentences from a source language to a target language, but to also comprehend the financial contextual meaning in different languages. \textbf{MINDS-14} \cite{gerz2021multilingual} consists of 8,168 samples of banking voice assistant data in text and audio formats across 14 different languages. \textbf{MultiFin} \cite{jorgensen2023multifin} includes 10,048 samples covering financial topics with 6 high-level labels (e.g., Finance) and 23 low-level labels (e.g., M\&A \& Valuations), sourced from public financial articles in 15 different languages.

\textbf{Market Forecasting} (MF) is an essential task in financial markets, involving the prediction of market price, volatility, and risk. This task extends beyond Stock Movement Prediction (SMP), which formulates problems as a classification task. The datasets introduced in Sentiment Analysis, Event Detection, and Multimodal tasks are also designed for Market Forecasting. Here, we include a list of datasets relevant to MF: \textbf{StockEmotions} (SA) \cite{lee2023stockemotions}, \textbf{EDT} (ED) \cite{zhou2021trade}, \textbf{MAEC} (MM-audio) \cite{li2020maec}, and \textbf{MONOPOLY} (MM-video) \cite{mathur2022monopoly}.

\section{Opportunities and Challenges}
In this section, we highlight various aspects guiding the future directions of FinLLMs, covering datasets, techniques, evaluation, implementation, and real-world applications.

\textbf{Datasets :}
High-quality data and multimodal data are significantly important for developing sophisticated FinLLMs. As most FinLLMs train general-domain LLMs on financial-specific data, the challenge lies in collecting high-quality financial data in diverse formats. Building \textbf{instruction fine-tuned financial datasets} by converting existing datasets for specific financial NLP tasks will facilitate the development of advanced FinLLMs. 
Also, the research on \textbf{financial multimodal datasets} will become increasingly important, enhancing the performance of FinLLMs on complex tasks.


\textbf{Techniques :}
Major challenges in finance include utilizing internal data without privacy breaches, causing security issues, while also enhancing trust in the responses generated by FinLLMs. To address these challenges, some actively researched techniques on LLMs, such as \textbf{Retrieval Augmented Generation} (RAG) \cite{lewis2020retrieval}, can be implemented in the financial domain. The RAG system is similar to an open-book approach, which retrieves non-pre-trained external knowledge resources (e.g., queried private data) to enhance the pre-trained model's raw representation of information.
RAG provides the model with access to factual information, enabling the generation of cross-referenced answers, therefore improving \textbf{reliability}, and minimizing \textbf{hallucination} issues. Moreover, RAG enables the use of internal non-trainable data without retraining the entire model, ensuring privacy is not breached.

\textbf{Evaluation :}
The primary challenge in evaluation is incorporating \textbf{domain knowledge from financial experts} to validate the model's performance based on financial NLP tasks. The current evaluation results were presented using commonly used NLP metrics such as F1-score or Accuracy. However, the knowledge-driven tasks require human evaluation by financial experts, appropriate financial evaluation metrics over NLP metrics, and expert feedback for model alignment. 
Furthermore, \textbf{advanced financial NLP tasks}, including the eight further benchmarks we presented, would discover the hidden capabilities of FinLLMs. These complex tasks will assess whether FinLLMs can serve as general financial problem-solver models \cite{guo2023chatgpt}, considering both cost and performance for specific tasks.

\textbf{Implementation :}
The challenge in selecting suitable FinLLMs and techniques lies in the \textbf{trade-off between cost and performance}. Depending on the task complexity and inference cost, selecting general-domain LLMs with prompting or task-specific models might be a more practical choice than building FinLLMs. This requires \textbf{LLMOps engineering} skills, including soft prompt techniques such as Parameter-Efficient Fine-Tuning (PEFT) and monitoring operation systems with a Continuous Integration (CI) and Continuous Delivery (CD) pipeline. 

\textbf{Applications :}
The challenge of developing real-world financial applications relates to non-technical issues, including business needs, industry barriers, data privacy, accountability, ethics, and the understanding gap between financial experts and AI experts. To overcome these challenges, sharing FinLLM \textbf{use-cases} will be beneficial across various financial fields including robo-advisor, quantitative trading, and low-code development \cite{yang2023fingpt}. 
Furthermore, we encourage future directions towards \textbf{generative applications} including report generation and document understanding. 

\section{Conclusion}

Our survey provides a concise yet comprehensive investigation of FinLLMs, by exploring their evolution from general-domain LMs, comparing techniques of FinPLMs/FinLLMs, and presenting six conventional benchmarks as well as eight advanced benchmarks and datasets. For future research, our big-picture view of FinLLMs, a relevant and extensive collection of datasets for more advanced evaluation, and opportunities and challenges for new directions for advanced FinLLMs will be beneficial to both the Computer Science and Finance research communities.



\appendix



\bibliographystyle{named}
\bibliography{ijcai24}

\end{document}